\documentclass{article}
\usepackage{natbib}
\usepackage{subcaption}

\usepackage{lscape}
\usepackage{adjustbox}
\usepackage[toc,page]{appendix}
\usepackage{arxiv}

\usepackage[utf8]{inputenc} % allow utf-8 input
\usepackage[T1]{fontenc}    % use 8-bit T1 fonts
\usepackage{hyperref}       % hyperlinks
\usepackage{url}            % simple URL typesetting
\usepackage{booktabs}       % professional-quality tables
\usepackage{amsfonts} 
\usepackage{amsmath}% blackboard math symbols
\usepackage{nicefrac}       % compact symbols for 1/2, etc.
\usepackage{microtype}      % microtypography
\usepackage{lipsum}
\usepackage{graphicx}
\graphicspath{ {./images/} }
\usepackage{fancyhdr}
\fancypagestyle{firstpage}{%
    \fancyhf{}
    \fancyfoot[L]{\footnotesize{*Correspondence to \href{mailto:gespitia3@utexas.edu}{gespitia3@utexas.edu}\\
    Preprint. Under review.}}

}

\title{Protein Structure Prediction in the 3D HP Model
Using Deep Reinforcement Learning}

\author{
 Giovanny Espitia* \\
  Department of Physics\\
  The University of Texas at Austin\\
  Austin, TX 78712 \\
  \\
   \And
 Yui Tik Pang \\
  School of Physics\\
  Georgia Institute of Technology\\
  Atlanta, GA 30332 \\
  \And
 James C. Gumbart \\
  School of Physics\\
  Georgia Institute of Technology\\
  Atlanta, GA 30332 \\
\\
} % This removes the arxiv preprint header and date
\date{}

\begin{document}
\maketitle
\thispagestyle{firstpage}

\begin{abstract}
We address protein structure prediction in the 3D Hydrophobic-Polar lattice model through two novel deep learning architectures. For proteins under 36 residues, our hybrid reservoir-based model combines fixed random projections with trainable deep layers, achieving optimal conformations with 25\% fewer training episodes. For longer sequences, we employ a long short-term memory network with multi-headed attention, matching best-known energy values. Both architectures leverage a stabilized Deep Q-Learning framework with experience replay and target networks, demonstrating consistent achievement of optimal conformations while significantly improving training efficiency compared to existing methods.
\end{abstract}

% keywords can be removed

\section{Introduction and Related Work}
Simulating protein folding is a fundamental challenge in biophysics and computational biology, yet it is crucial for understanding protein structure, function, and dynamics, with significant implications for drug discovery and disease diagnosis. The Hydrophobic-Polar (HP) model serves as a simplified yet powerful framework for studying protein folding, classifying amino acids as either hydrophobic (H) or polar (P) on a lattice structure. Despite its apparent simplicity, finding optimal conformations in the HP model remains NP-complete, making it particularly challenging for larger proteins. Early approaches to this problem employed various computational methods, including genetic algorithms \citep{unger-1993}, Monte Carlo simulations with evolutionary algorithms \citep{patton-1995}, and memetic algorithms with self-adaptive local search \citep{n-2010}. Additional methodologies encompassed ant colony optimization \citep{shmygelska-2005}, core-directed chain growth \citep{beutler-1996}, and the pruned-enriched Rosenbluth method (PERM) \citep{grassberger-1997}.
Recent advances in deep reinforcement learning (DRL) have opened new avenues for addressing the protein folding challenge. Notable contributions include Q-learning approaches \citep{czibula-2011}, hybrid methods combining Q-learning with ant colony optimization \citep{dogan-2015}, and FoldingZero \citep{li-2018}, which integrates Monte Carlo tree search with convolutional neural networks. Building upon these foundations, we propose two novel architectures: (1) a reservoir computing-based hybrid architecture \citep{jaeger-2007} that captures temporal dependencies in the protein folding process and (2) an LSTM network enhanced with multi-head attention layers that effectively models long-range interactions between amino acids \citep{bahdanau-2015}. These architectural approaches have individually demonstrated success in various domains, including time series prediction \citep{subramoney-2021}, speech recognition \citep{araujo-2020}, and robot control \citep{antonelo-2015}.Our work represents the first application of reservoir computing to the protein folding problem in the 3D HP model, while also introducing an attention-enhanced LSTM architecture specifically designed for longer protein sequences. The reservoir-based approach leverages the computational efficiency of fixed, randomly initialized recurrent neural networks to project input data into high-dimensional space, while the attention mechanism in the LSTM architecture enables effective modeling of interactions between distant amino acids. Both architectures consistently achieve optimal conformations matching the best known energy values while demonstrating improved training efficiency compared to traditional approaches.
The remainder of this paper is organized as follows: Section 2 formulates the protein folding problem in the 3D HP model as a deep reinforcement learning problem and describes our proposed architectures and training methodology, Section 3 presents experimental results comparing our approach with other state-of-the-art methods and analyzing the role of both the reservoir and attention mechanisms, and Sections 4 and 5 discuss implications and conclude the paper. 

\section{Methodology}

In this section, we describe the methods used to model the problem as Markov Decision Process (MDP) and introduce the details of the architectures. 

\subsection{Modeling the problem in a cubic lattice}
Let $\mathcal{G}$ be a 3D cubic lattice with lattice points $(x, y, z) \in \mathbb{Z}^3$. Similar to \citep{yang-2023}, the protein folding process is modeled as a self-avoiding walk (SAW) on $\mathcal{G}$, where each amino acid in the sequence occupies a single lattice point, and no two amino acids can occupy the same point simultaneously. The SAW begins by placing the first two amino acids at positions $\vec{r}_0 = (0, 0, 0)$ and $\vec{r}_1 = (0, 1, 0)$ in the cubic lattice. The following constraints are imposed on the placement of subsequent amino acids:

\begin{itemize}

\item \textbf{Distance constraint:} The distance between consecutive amino acids in the sequence must be exactly one lattice unit. Let $\vec{r}_i = (x_i, y_i, z_i)$ and $\vec{r}_{i+1} = (x_{i+1}, y_{i+1}, z_{i+1})$ be the positions of two consecutive amino acids. Then, the distance constraint can be expressed as:
\begin{equation}
\begin{aligned}
\lVert \vec{r}_{i+1} - \vec{r}_i \rVert &= \left((x_{i+1} - x_i)^2 + (y_{i+1} - y_i)^2 \right. \\
&\quad \left. + (z_{i+1} - z_i)^2 \right)^{\frac{1}{2}} = 1
\end{aligned}
\end{equation}

\item \textbf{Bond angle constraint:} The bond angles formed by three consecutive amino acids must be restricted to 90° or 180° to ensure rotational invariance. Let $\vec{r}_i$, $\vec{r}_{i+1}$, and $\vec{r}_{i+2}$ be the positions of three consecutive amino acids. The bond angle constraint can be expressed as:
\begin{equation}
\begin{aligned}
   (\vec{r}_{i+1} - \vec{r}_i) \cdot (\vec{r}_{i+2} - \vec{r}_{i+1}) = 0 \quad & \;\text{(for a 90° angle)} \\ 
   (\vec{r}_{i+1} - \vec{r}_i) \times (\vec{r}_{i+2} - \vec{r}_{i+1}) = \vec{0} & \;\;\;\; \text{(for a 180° angle)}
\end{aligned}
\end{equation}
where $\cdot$ denotes the dot product and $\times$ denotes the cross product.

\item \textbf{Self-avoidance constraint:} The chain must not intersect with itself, i.e., no two amino acids can occupy the same lattice point. This can be expressed as:
\begin{equation}
\vec{r}_i \neq \vec{r}_j \quad \forall i, j \in {0, 1, \dots, N-1}, i \neq j
\end{equation}
where $N$ is the total number of amino acids in the sequence.
\item \textbf{Translational invariance:} Translational invariance is ensured by the presence of primitive translation vectors $(\vec{a}_1, \vec{a}_2, \vec{a}_3)$ that map the lattice onto itself. The translation vectors are defined as:
\begin{equation}
\vec{T} = n_1\vec{a}_1 + n_2\vec{a}_2 + n_3\vec{a}_3
\end{equation}
where $n_1, n_2, n_3 \in \mathbb{Z}$.
\end{itemize}

In this problem, the objective is to find the optimal conformation that maximizes the number of hydrophobic-hydrophobic (H-H) contacts in the folded protein. An H-H contact occurs when two non-consecutive hydrophobic amino acids are placed adjacent to each other in the lattice. The energy of a given fold is defined as the negative of the total number of valid H-H contacts:
\begin{equation}
E = -(\text{number of valid H-H contacts})
\end{equation}
By minimizing the energy function, the most stable conformation with the maximum number of H-H contacts can be found.

\subsection{DRL Setup}

By treating the protein folding process as a SAW, the RL agent places the amino acids sequentially. In our setup, the protein is considered the agent, and the cubic lattice represents the environment. The length of the path corresponds to the number of amino acids in the protein sequence. At the end of each episode, the environment provides the agent with a reward based on the energy of the achieved fold.

\subsubsection{Markov Decision Process Formulation}
A MDP is defined by a tuple $(\mathcal{S}, \mathcal{A}, \mathcal{P}, \mathcal{R}, \gamma)$, where $\mathcal{S}$ is the state space, $\mathcal{A}$ is the action space, $\mathcal{P}$ is the transition probability function, $\mathcal{R}$ is the reward function, and $\gamma \in [0, 1]$ is the discount factor. At each discrete time step $t$, the agent interacts with the environment by observing the current state $s_t \in \mathcal{S}$ and taking an action $a_t \in \mathcal{A}$. The environment then transitions to a new state $s_{t+1} \in \mathcal{S}$ according to the transition probability function $\mathcal{P}(s_{t+1} | s_t, a_t)$ and provides a reward $r_{t+1} \in \mathcal{R}$ to the agent. The objective of the agent is to learn a policy $\pi: \mathcal{S} \rightarrow \mathcal{A}$ that maximizes the expected cumulative reward over an episode, defined as $\mathbb{E}\left[\sum_{t=0}^T \gamma^t r_t\right]$, where $T$ is the length of the episode.

\subsubsection{Deep Q-Learning with Stabilization Techniques}
Deep Q-learning (DQN) is a reinforcement learning algorithm that combines Q-learning with deep neural networks to optimize the conformations achieved by the agent in the protein folding problem. Q-learning is a value-based method that learns the optimal action-value function, or Q-function, which represents the expected future reward for taking a particular action in a given state. The Q-function, denoted as $Q(s, a)$, satisfies the Bellman optimality equation:
\begin{equation}
Q(s, a) = \mathbb{E}\Big[r_{t+1} + \gamma \max_{a_{t+1}} Q(s_{t+1}, a_{t+1})\Big]
\end{equation}
where $r_{t+1}$ is the reward received at time step $t+1$ for carrying out action $a_{t}$ at state $s_{t}$, $\gamma \in [0, 1]$ is the discount factor that determines the importance of future rewards, and $s_{t+1}$ is the state at time step $t+1$. The expectation $\mathbb{E}$ is taken over all possible next states and actions, reflecting the average outcome based on the agent's policy and the environment's dynamics. The optimal Q-function, denoted as $Q^*(s, a)$, satisfies the Bellman optimality equation and provides the maximum expected future reward for taking action $a$ in state $s$. DQN addresses the limitations of traditional Q-learning, which becomes intractable for problems with large state spaces, by approximating the Q-function using a deep neural network. The neural network, parameterized by $\theta$, takes the state $s$ as input and outputs the Q-values for each action $a$. The training objective is to minimize the mean-squared error loss function:
\begin{equation}
\mathcal{L}(\theta) = \mathbb{E}\Big[\big(r + \gamma \max_{a_{t+1}} Q(s_{t+1}, a_{t+1}; \theta^-) \\
- Q(s, a; \theta)\big)^2\Big]
\end{equation}
where $s_{t+1}$ is the next state, $a_{t+1}$ is the next action, and $\theta^-$ represents the parameters of a target network. The target network is a separate neural network that is periodically updated with the parameters of the main network, denoted as $\theta$. The use of a target network is a key stabilization technique introduced in the DQN paper by \citep{mnih-2015}, which helps to mitigate the issue of divergence in the learning process.
During training, the agent interacts with the environment and stores the experienced transitions $e_{t} = (s_t, a_t, r_{t+1}, s_{t+1})$ in a replay buffer $\mathcal{D}$. The replay buffer is a fixed-size cache that stores the most recent transitions experienced by the agent. At each training step, a minibatch of transitions ${e_1, e_2, \dots, e_N}$ is sampled uniformly from the replay buffer to update the parameters of the main network $\theta$. The use of a replay buffer helps to break the correlation between consecutive samples and stabilizes the learning process by providing a diverse set of experiences for training.

\begin{figure}[ht]
\vskip 0.2in
\begin{center}
\centerline{\includegraphics[scale=0.5]{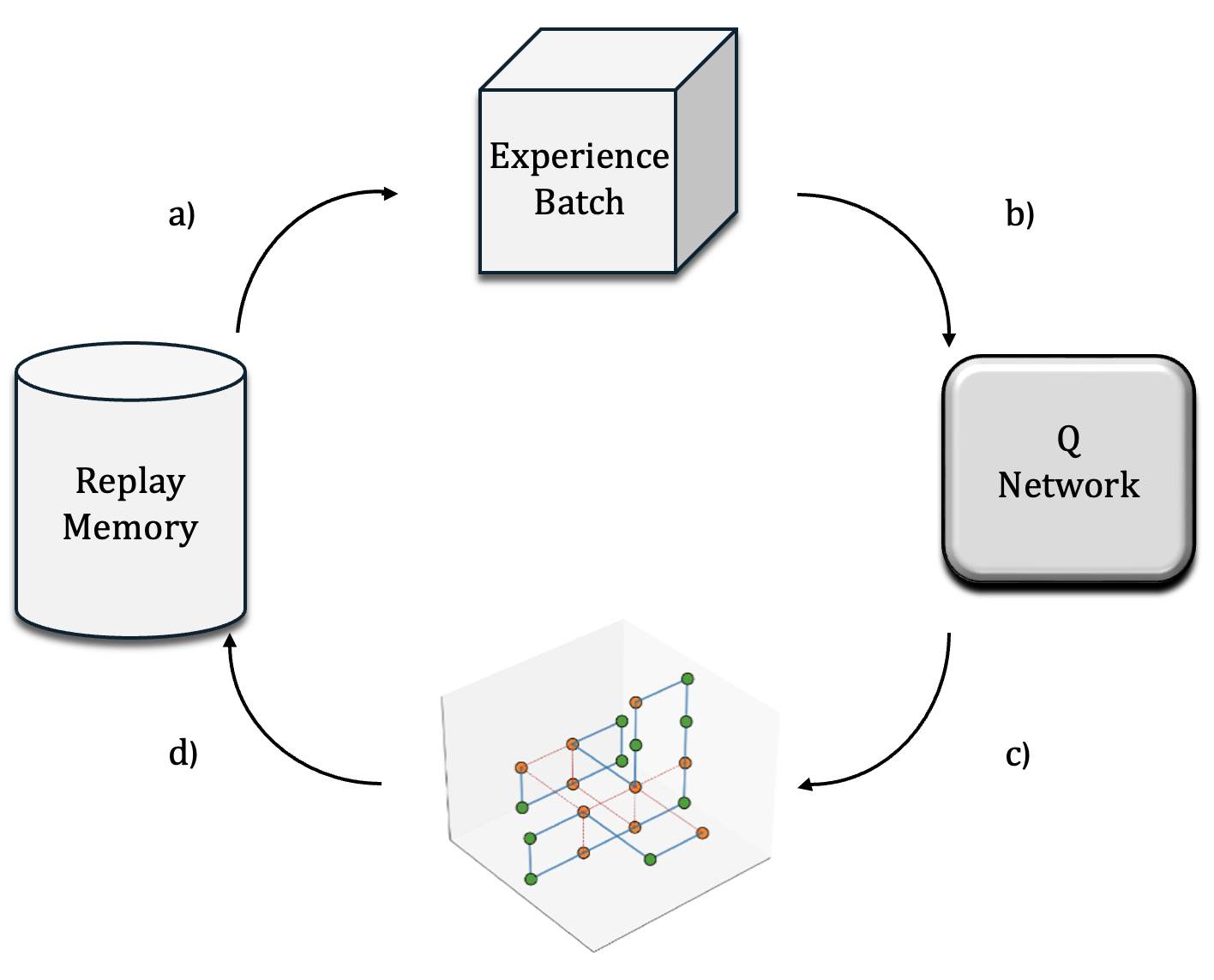}}
\caption{Deep reinforcement learning training loop. In a), we sample a batch of experience from the buffer. The batch then serves as input to the Q - network in b). Based on the output Q - value tensor, the agent makes a decision in c) to take $a_{t+1}$ that corresponds to the greatest value of the Q - output tensor. In step d), the experience is stored in the replay memory..}
\label{training_process}
\end{center}
\vskip -0.2in
\end{figure}

The training process is displayed in Figure 1 and proceeds as follows:

\begin{enumerate}
\item Sample a minibatch of transitions ${e_1, e_2, \dots, e_N}$ from the replay buffer $\mathcal{D}$.
\item For each transition $e_t = (s_t, a_t, r_{t+1}, s_{t+1})$, compute the target:
\begin{equation}
y_t = \begin{cases}
r_{t+1} \;\;\; \text{if } s_{t+1} \text{ is terminal} \\
r_{t+1} + \gamma \max_{a'} Q(s_{t+1}, a'; \theta^-) & \text{otherwise}
\end{cases}
\end{equation}
\item Update the parameters $\theta$ of the main network by minimizing the loss function:
\begin{equation}
\mathcal{L}(\theta) = \frac{1}{N} \sum_{t=1}^N \left(y_t - Q(s_t, a_t; \theta)\right)^2
\end{equation}
\item Every $C$ steps, update the parameters $\theta^-$ of the target network by copying the parameters $\theta$ of the main network:
\begin{equation}
\theta^- \leftarrow \theta
\end{equation}
\item Select an action $a_t$ based on the current state $s_t$ using an $\epsilon$-greedy policy derived from the Q-values:
\begin{equation}
a_t = \begin{cases}
\text{argmax}\;Q(s_t, a; \theta) & \text{with probability } 1 - \epsilon \\
\text{random action} & \text{with probability } \epsilon
\end{cases}
\end{equation}
where $\epsilon$ is the exploration rate that determines the probability of selecting a random action instead of the greedy action with the highest Q-value. 
\item Execute the selected action $a_t$ in the environment and observe the next state $s_{t+1}$ and reward $r_{t+1}$.
\item Store the transition $(s_t, a_t, r_{t+1}, s_{t+1})$ in the replay buffer $\mathcal{D}$.
\end{enumerate}

\subsubsection{State Representation}
We represent the states using one-hot encoded vectors. Each state vector consists of two parts: the first part represents the available actions (forward (F), backward (B), right (R), left (L), up (U), and down (D)), and the second part encodes the type of amino acid (H or P) at the current position. Specifically, the state is an 8-dimensional vector, where the first six elements correspond to the possible actions the agent can take: No Decision (ND), F, L, R, U, D. The last two elements represent the type of amino acid (H or P).
For each training episode, we obtain a 3D shape tensor ($N$,8,1), where $N$ is the length of the protein sequence. The first dimension corresponds to the time steps, that is, the positions in the protein sequence, while the second dimension represents the state features (actions and amino acid type). The third dimension is a singleton dimension to facilitate input to the neural network. This state representation allows the agent to make informed decisions based on the available actions and the type of amino acid at each position. By concatenating action and amino acid information, the neural network can learn the dependencies between amino acid placements and the resulting energy conformations. It is important to note that the neural network does not directly observe amino acid placements; instead, it deduces this information from actions taken thus far. The ``No Decision'' (ND) indicates a state where no further moves are possible, leading to episode termination. The backward action (B) is excluded because once an amino acid is placed at a position, revisiting that position is not feasible; thus, it cannot be performed in this context.

\subsection{Q - Network Architectures}

In this subsection, we describe the two architectures proposed in this paper.

\subsubsection{Hybrid - Reservoir}
The architecture consists of a reservoir layer followed by several fully connected layers, as illustrated in Figure 2.

\begin{figure}[ht]
\vskip 0.2in
\begin{center}
\includegraphics[scale=0.45]{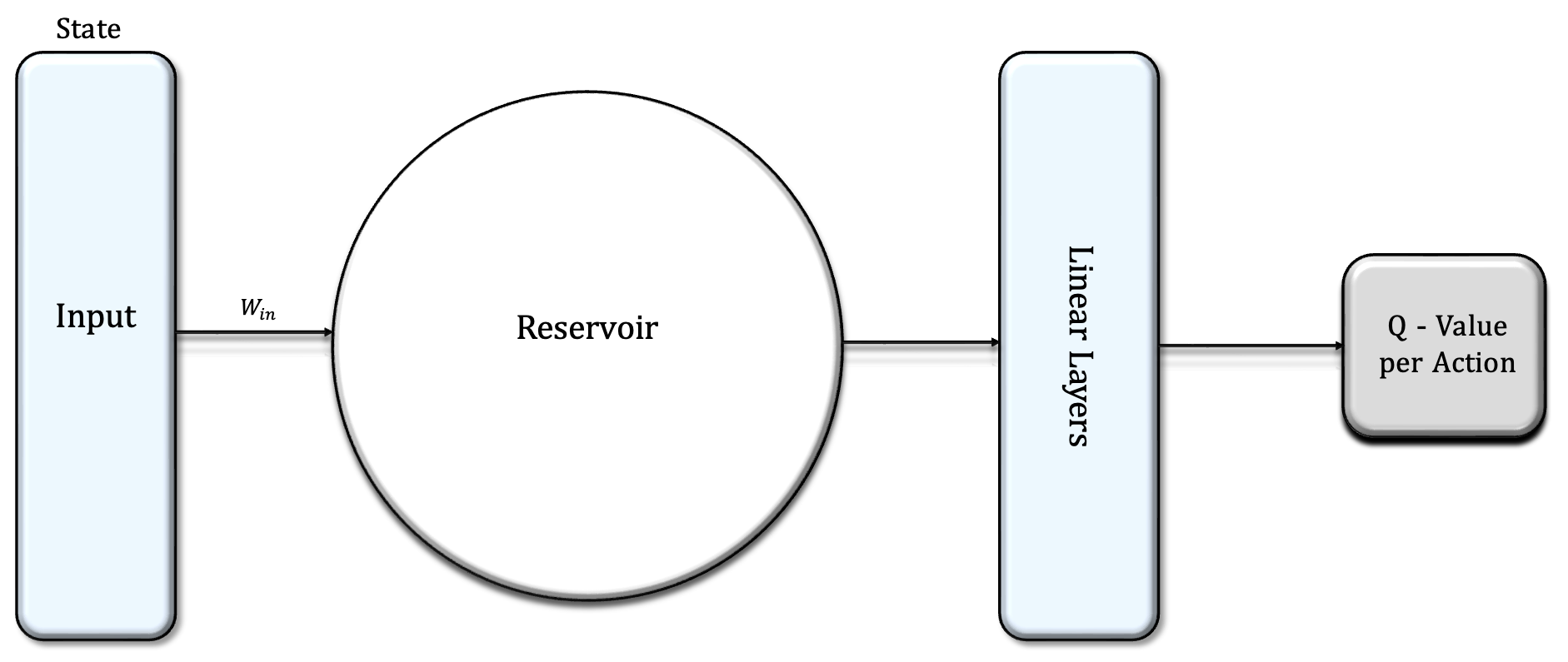}
\caption{The input layer consists of a (N, 8, 1) tensor representing the state at a particular timestep. The reservoir is a randomly initialized weight matrix with a topology specified beforehand. The linear layers consists of a simple fully-connected feed forward neural network. The output is a (5, 1) tensor representing the Q - value or future expected total reward per action.}
\label{training_process}
\end{center}
\vskip -0.2in
\end{figure}

The input to the reservoir neural network is a flattened vector representation of the state, denoted as $\mathbf{x} \in \mathbb{R}^{N \times d}$, where $N$ is the sequence length and $d$ is the dimensionality of the one-hot encoded vector described in 2.2.3. The reservoir layer applies a fixed random projection of the input into a high-dimensional space. Mathematically, the reservoir layer can be described as follows:
\begin{equation}
\mathbf{r}(t) = f(\mathbf{W}_{\rm in} \mathbf{x}(t) + \mathbf{W} \mathbf{r}(t-1))
\end{equation}
where $\mathbf{r}(t) \in \mathbb{R}^{N_r}$ is the reservoir state at time step $t$, $\mathbf{x}(t) \in \mathbb{R}^{d}$ is the input state, $\mathbf{W}_{\rm in} \in \mathbb{R}^{N_r \times d}$ is the trainable input weight matrix, $\mathbf{W} \in \mathbb{R}^{N_r \times N_r}$ is the reservoir weight matrix, and $f(\cdot)$ is the activation function (e.g., hyperbolic tangent). The reservoir weight matrix $\mathbf{W}$ is randomly initialized and remains fixed during training. It follows a specific connectivity pattern, such as the Erd\H{o}s-R\'{e}nyi topology, which promotes a limited number of active connections among neurons. The sparsity and connectivity of the reservoir are important factors in determining its computational capacity and ability to capture complex dynamics. The size of the reservoir, denoted as $N_r$, is a hyperparameter that depends on the length and complexity of the protein sequence. Empirically, we found that a reservoir size of 1000 works well for sequences of length $N \leq 36$, while for longer sequences ($N = 48, 50$), a larger reservoir size of around 3000 is used to capture the increased complexity. The reservoir layer applies a hyperbolic tangent (tanh) activation function to introduce nonlinearity. The output of the reservoir layer is then passed through a series of fully connected layers with Rectified Linear Unit (ReLU) activation functions. The sizes of the fully connected layers are 512, 256, 128, and 84, respectively. These layers learn to extract meaningful features from the reservoir representation and progressively reduce the dimensionality of the feature space. The final output layer is a fully connected layer with a size equal to the number of actions, which generates the Q-values for each action.

\subsubsection{LSTM with Multi-Head-Attention}

For long proteins, we employ an LSTM architecture with multi-head attention \citep{vaswani-2017}. The architecture consists of an LSTM layer, a multi-head attention mechanism, and a fully connected output layer, as illustrated in Figure 3. 

\begin{figure}[ht]
\vskip 0.2in
\begin{center}
\includegraphics[scale=0.37]{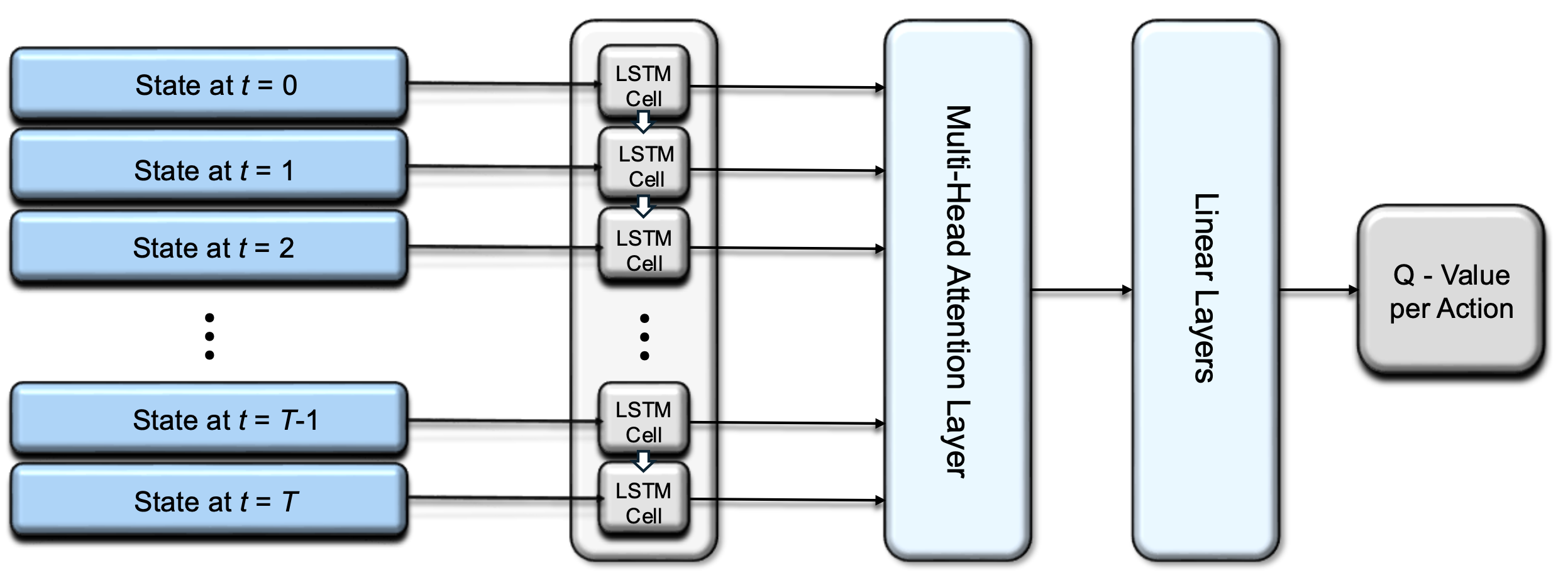}
\caption{LSTM-A architecture for protein folding. Sequential states are processed through LSTM cells, generating hidden states that are weighted by an 8-head attention mechanism. The attention output is mapped to action Q-values through a fully connected layer, enabling the model to leverage both sequential patterns and long-range dependencies.}
\label{training_process}
\end{center}
\vskip -0.2in
\end{figure}

The multi-head attention mechanism enhances the network's ability to focus on different aspects of the input sequence simultaneously. Given an input sequence processed by the LSTM layers producing hidden states \(\mathbf{H} \in \mathbb{R}^{N \times d}\), where \(N\) is the sequence length and \(d\) is the hidden dimension, the attention mechanism computes attention patterns across four different representation subspaces. For each attention head \(i\), the mechanism computes:

\begin{equation}
\text{Attention}_i(\mathbf{Q}, \mathbf{K}, \mathbf{V}) = \text{softmax}\left(\frac{\mathbf{Q}\mathbf{K}^T}{\sqrt{d_k}}\right)\mathbf{V}
\end{equation}

where \(\mathbf{Q}\), \(\mathbf{K}\), and \(\mathbf{V}\) are linear projections of the LSTM output \(\mathbf{H}\), and \(d_k\) is the dimension of the key vectors. The outputs from all heads are concatenated and projected to the required dimensionusing a weight matrix $\mathbf{W}^O$:

\begin{equation}
\text{MultiHead}(\mathbf{H}) = \text{Concat}(\text{head}_1,\ldots,\text{head}_4)\mathbf{W}^O
\end{equation}

Our implementation uses a hidden size of \(d=512\) with four attention heads. The LSTM layer processes the input sequence with batch-first ordering, and the attention mechanism operates on the full sequence of LSTM outputs. The final output is obtained by selecting the last time step of the attention output, followed by a fully connected layer that maps to the action space dimension.

\section{Experiments and Results}
To evaluate our approach, we utilized sequences with best-known energy values published in Table 1 from \citep{boumedine-2022}. Figure 4 illustrates the optimal folds for various sequences, providing a benchmark for our experiments. We began our investigation with the baseline Fully Connected Feedforward Neural Network (FFNN), which employs a four-layer architecture (512→256→84→output-size) with ReLU activations. This model was effective for shorter sequences (N $\le$ 36), demonstrating reasonable performance in achieving optimal energy states. However, it struggled with longer sequences, often failing to converge to the best-known values for more complex protein structures. This limitation highlighted the need for enhancements that could better capture the intricate dynamics of protein folding. To address these shortcomings, we introduced the Reservoir-based Fully Connected Feedforward Neural Network (FFNN-R). By incorporating a reservoir layer initialized using Xavier uniform initialization and tanh activation, FFNN-R enhanced the network’s temporal memory capabilities while maintaining training efficiency. The reservoir's sparsity and specific connectivity patterns allowed for improved processing of input data. Empirical results showed that FFNN-R outperformed the vanilla FFNN, converging to optimal energy states faster, requiring approximately 25\% fewer training episodes to reach these conformations. Particularly for shorter sequences, FFNN-R consistently achieved BKVs for sequences A1-A5 and A8-A10, showcasing its superior convergence speed. Building on the improvements seen with FFNN-R, we explored LSTM architectures to further enhance performance on longer sequences. The LSTM-OLH architecture utilized traditional LSTM cells while relying solely on the last hidden state for decision-making. While this approach provided some improvements over FFNN, it still struggled to effectively capture long-range dependencies critical in protein folding.
To maximize performance further, we developed the LSTM-A architecture, which incorporated a multi-head attention mechanism. This enhancement allowed the model to dynamically weigh different temporal aspects of the state sequence, effectively capturing long-range interactions between amino acids. The results indicated that LSTM-A significantly outperformed both LSTM-OLH and vanilla FFNN implementations for longer sequences (N > 36). It consistently achieved optimal or near-optimal energy states that matched BKVs for sequences 3d1-3d5 and approached BKVs for more challenging sequences like 3d6-3d9. Throughout our experiments, we employed a stabilized Deep Q-Learning framework with experience replay and an epsilon-greedy strategy for action selection. The training utilized a single NVIDIA H100 GPU with architectures implemented in PyTorch version 2.5.0 running on CUDA version 11.8. The Adam optimizer was used with a learning rate of 0.001, and we selected the smooth L1 loss function for its effectiveness in DQL tasks with discrete action spaces. As shown in Figure 5, for shorter sequences (N $\le$ 36), FFNN-R outperformed other architectures and converged to optimal energy states more rapidly. However, for longer sequences, LSTM-A exhibited superior performance compared to both LSTM-OLH and vanilla FFNN implementations. The training process revealed that the attention mechanism in LSTM-A began to play a crucial role after approximately 100,000 training episodes, coinciding with significant improvements in energy minimization. This suggests that the network first learns local folding patterns before leveraging attention to capture long-range dependencies.
In contrast, the FFNN-R architecture demonstrated rapid initial convergence, often reaching near-optimal conformations within the first 50,000 episodes, particularly for shorter sequences. Moreover, FFNN-R provided more efficient parameter utilization as it required only 264,445 trainable parameters compared to LSTM-A’s 6,324,741. In summary, our findings illustrate a clear progression in model performance from vanilla FFNN to reservoir-based architectures and finally to LSTM networks with attention mechanisms. Each architectural enhancement addressed specific challenges in protein folding prediction, culminating in models that not only achieved optimal conformations but also demonstrated improved training efficiency across varying sequence lengths.

\begin{figure}[ht]
\centering
\vspace{0.2in}
\begin{subfigure}[b]{0.5\columnwidth}
    \centering
    \includegraphics[scale=0.3]{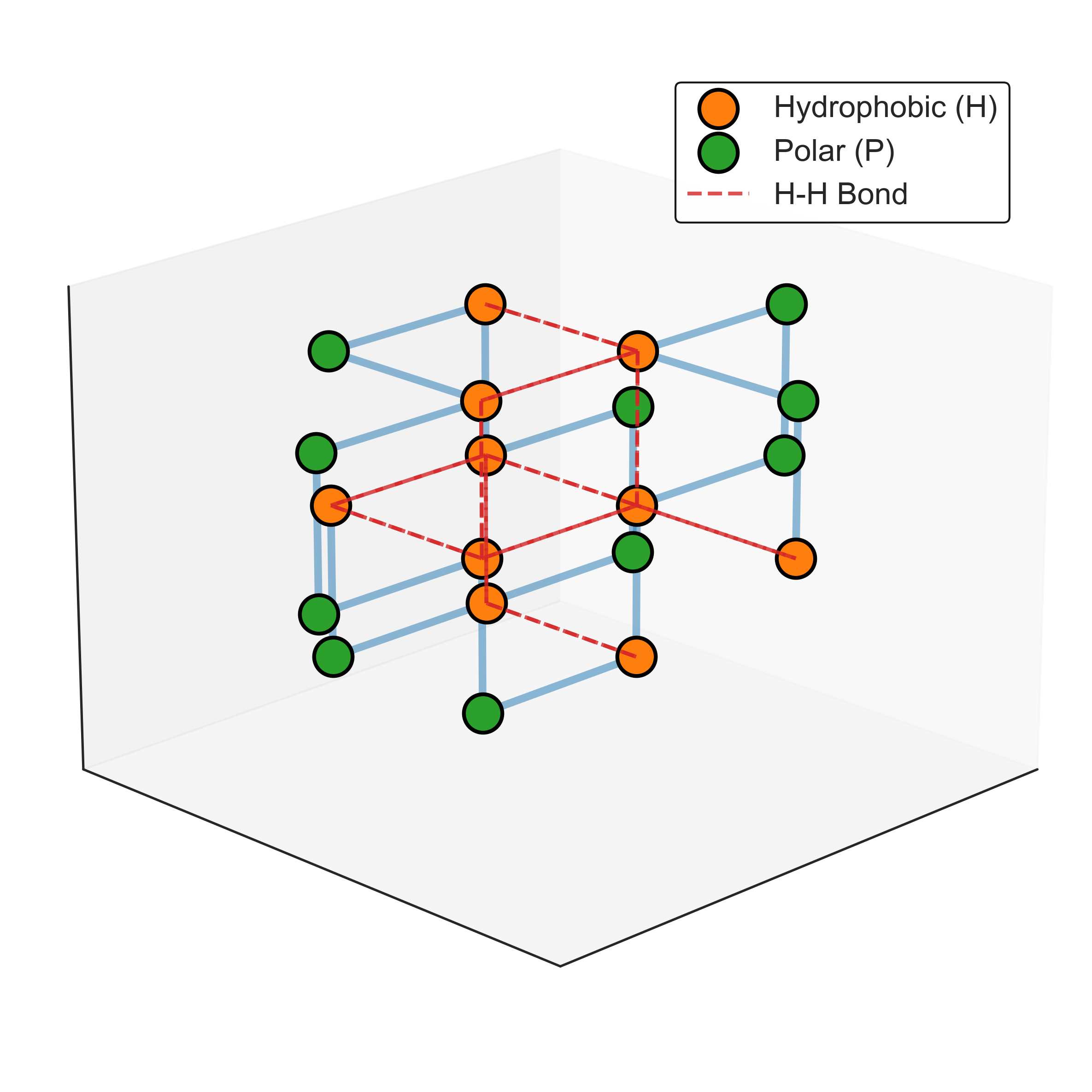}
    \caption{3d1}
    \label{fig:sub1}
\end{subfigure}%
\hfill
\begin{subfigure}[b]{0.5\columnwidth}
    \centering
    \includegraphics[scale=0.3]{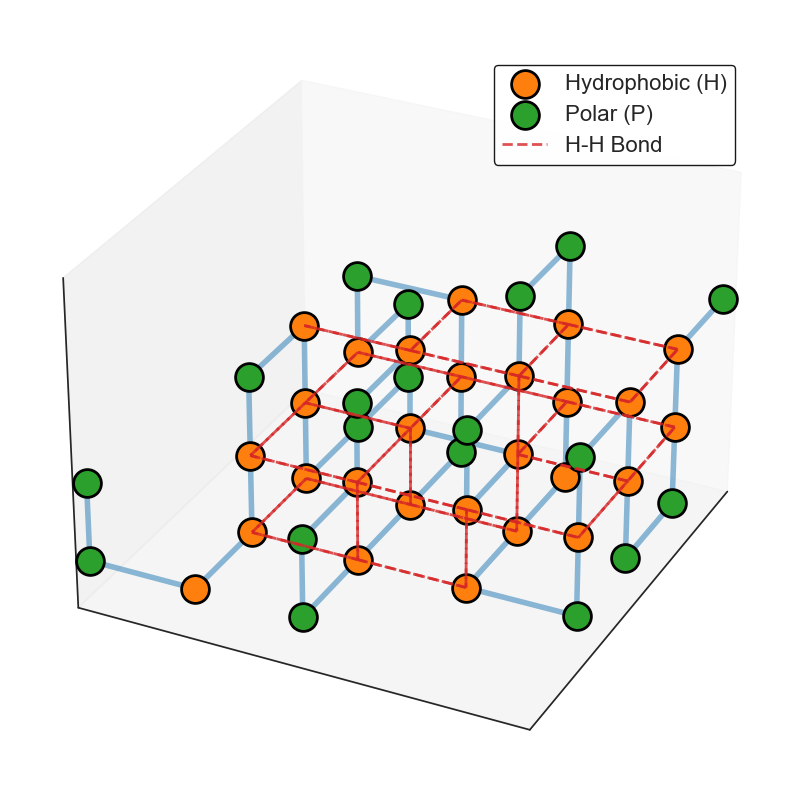}
    \caption{3d5}
    \label{fig:sub2}
\end{subfigure}

\vspace{0.1in}

\begin{subfigure}[b]{0.5\columnwidth}
    \centering
    \includegraphics[scale=0.3]{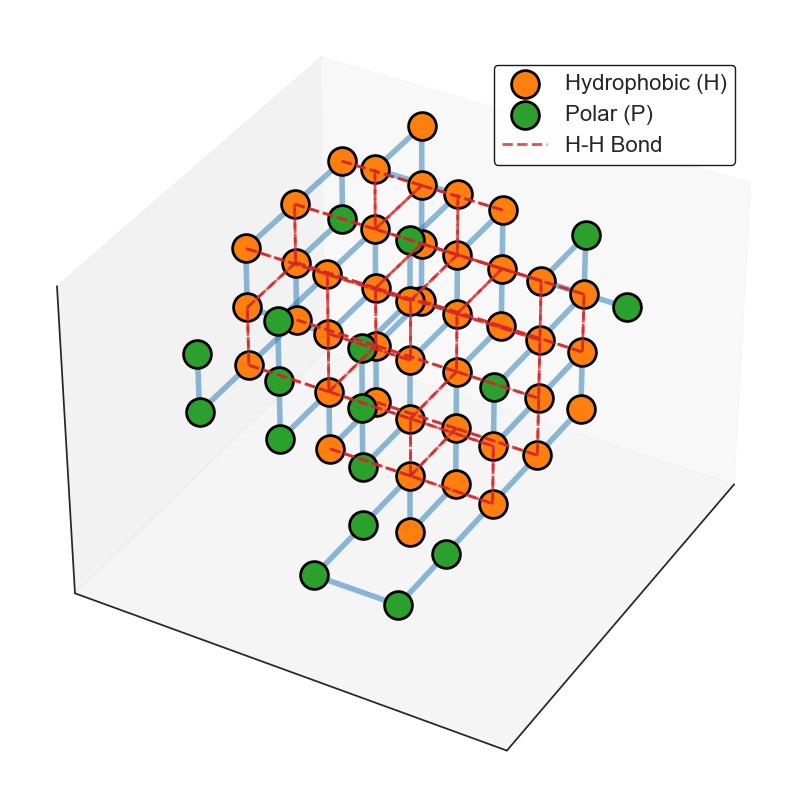}
    \caption{3d9}
    \label{fig:sub3}
\end{subfigure}%
\hfill
\begin{subfigure}[b]{0.5\columnwidth}
    \centering
    \includegraphics[scale=0.3]{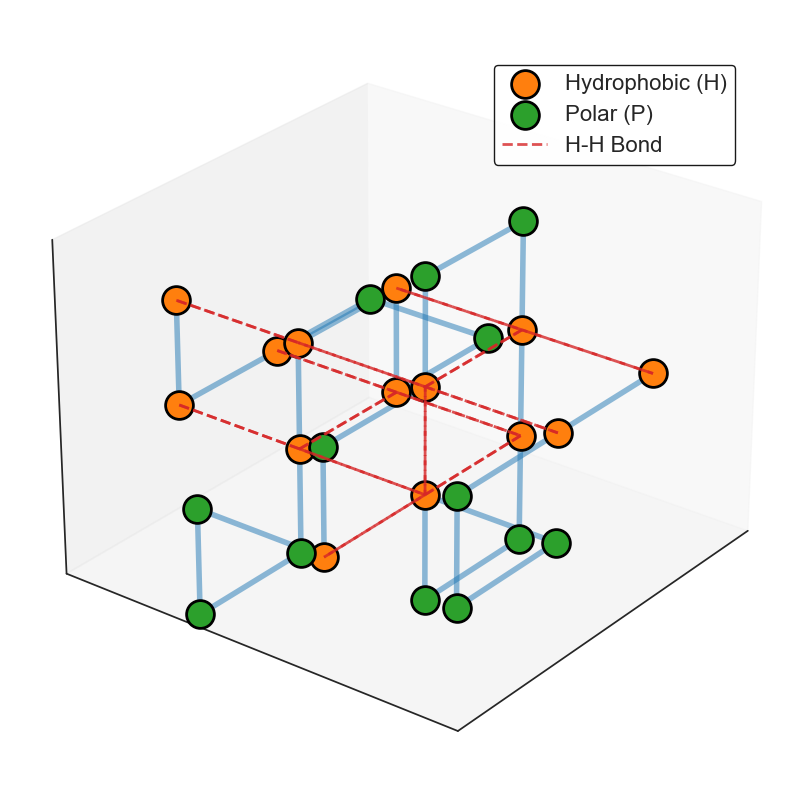}
    \caption{A4}
    \label{fig:sub4}
\end{subfigure}
\caption{Least Energy Conformations for different sequences.}
\label{fig:test}
\vspace{-0.2in}
\end{figure}

\begin{table}[t]
\caption{Sequences and their corresponding best known conformation energy values (BKV).}
\label{tab:my_label_table1}
\vskip 0.15in
\begin{center}
\begin{small}
\begin{sc}
\begin{tabular}{lcccccc}
\toprule
Seq. & Length & Sequence & BKV \\
\hline \\
3d1 & 20 & (HP)$^2$PH(HP)$^2$(PH)$^2$HP(PH)$^2$ & $-11$ \\
3d2 & 24 & H$^2$P$^2$(HP$^2$)$^6$H$^2$ & $-13$ \\
3d3 & 25 & P$^2$HP$^2$(H$^2$P$^4$)$^3$H$^2$ & $-9$ \\
3d4 & 36 & P(P$^2$H$^2$)$^2$P$^5$H$^5$(H$^2$P$^2$)$^2$P$^2$H(HP$^2$)$^2$ & $-18$ \\
3d5 & 46 & P$^2$H$^3$PH$^3$P$^3$HPH$^2$PH$^2$P$^2$HPH$^4$ & $-35$ \\
&    & PHP$^2$H$^5$PHPH$^2$P$^2$H$^2$P & \\
3d6 & 48 & P$^2$H(P$^2$H$^2$)$^2$P$^5$H$^{10}$P$^6$ & $-31$ \\
&    & (H$^2$P$^2$)$^2$HP$^2$H$^5$ & \\
3d7 & 50 & H$^2$(PH)$^3$PH$^4$PH(P$^3$H)$^2$P$^4$ & $-34$ \\
&    & (HP$^3$)$^2$HPH$^4$(PH)$^3$PH$^2$ & \\
3d8 & 58 & PH(PH$^3$)$^2$P(PH$^2$PH)$^2$H(HP)$^3$ & $-44$ \\
&    & (H$^2$P$^2$H)$^2$PHP$^4$(H(P$^2$H)$^2$)$^2$ & \\
3d9 & 60 & P(PH$^3$)$^3$H$^5$P$^3$H$^{10}$PHP$^3$ & $-55$ \\
&    & H$^{12}$P$^4$H$^6$PH$^2$PH & \\
\bottomrule
\end{tabular}
\end{sc}
\end{small}
\end{center}
\vskip -0.1in
\end{table}
\begin{table}[t]
\caption{Sequences and their corresponding best known conformation energy values (BKV) (continued).}
\label{tab:my_label_table2}
\vskip 0.15in
\begin{center}
\begin{small}
\begin{sc}
\begin{tabular}{lcccccc}
\toprule
Seq. & Length & Sequence & BKV \\
\hline \\
A$_1$ & 27 & PHPHPH$^3$P$^2$HPHP$^{11}$H$^2$P & $-9$ \\
A$_2$ & 27 & PH$^2$P$^{10}$H$^2$P$^2$H$^2$P$^2$HP$^2$HPH & $-10$ \\
A$_3$ & 27 & H$^4$P$^5$HP$^4$H$^3$P$^9$H & $-8$ \\
A$_4$ & 27 & H$^3$P$^2$H$^4$P$^3$HPHP$^2$H$^2$P$^2$HP$^3$H$^2$ & $-15$ \\
A$_5$ & 27 & H$^4$P$^4$HPH$^2$P$^3$H$^2$P$^{10}$ & $-8$ \\
A$_6$ & 27 & HP$^6$HPH$^3$P$^2$H$^2$P$^3$HP$^4$HPH & $-12$ \\
A$_7$ & 27 & HP$^2$HPH$^2$P$^3$HP$^5$HPH$^2$PHPHPH$^2$ & $-13$ \\
A$_8$ & 27 & HP$^{11}$HPHP$^8$HPH$^2$ & $-4$ \\
A$_9$ & 27 & P$^7$H$^3$P$^3$HPH$^2$P$^3$HP$^2$HP$^3$ & $-7$ \\
A$_{10}$ & 27 & P$^5$H$^2$PHPHPHPHP$^2$H$^2$PH$^2$PHP$^3$ & $-11$ \\
A$_{11}$ & 27 & HP$^4$H$^4$P$^2$HPHPH$^3$PHP$^2$H$^2$P$^2$H & $-16$ \\
\bottomrule
\end{tabular}
\end{sc}
\end{small}
\end{center}
\vskip -0.1in
\end{table}

\begin{table}[t]
\caption{Energy values of 3d protein sequences obtained by different algorithms. The right-most column corresponds to the least energy conformation found by our models. The results of all experiments tabulated in Table 4 in the appendix section.}
\label{tab:my_label_table1}
\begin{center}
\begin{small}
\begin{sc}
\begin{tabular}{lcccccc}
\toprule
Seq. & Length & \bf{BKV} & GA & BILS & LSTM-A\\
\midrule
3d1 & 20 & \bf{-11} & -11 & -10  & -11 \\
3d2 & 24 & \bf{-13} & -13 & -9  & -13 \\
3d3 & 25 & \bf{-9} & -9 & -7 &  -9 \\
3d4 & 36 & \bf{-18} & -18 & -12 & -18 \\
3d5 & 46 & \bf{-35} & -32 & -22 &  -33 \\
3d6 & 48 & \bf{-31} & -31 & -19 &  -30 \\
3d7 & 50 & \bf{-34} & -30 & -18 &  -32 \\
3d8 & 58 & \bf{-44} & -37 & -23 &  -40 \\
3d9 & 60 & \bf{-55} & -50 & -36 &  -51 \\
\midrule
Seq. & Length & BKV & GA & TPPSO  & FFNN-R\\
\midrule
$A_{1}$ & 27 & \bf{-9} & -8 & -9 &  -9 \\
$A_{2}$ & 27 & \bf{-10} & -10 & -10 &  -10 \\
$A_{3}$ & 27 & \bf{-8} & NA & -8 &  -8 \\
$A_{4}$ & 27 & \bf{-15} & -15 & -15 &  -15 \\
$A_{5}$ & 27 & \bf{-8} & -8 & -8 &  -8 \\
$A_{6}$ & 27 & \bf{-12} & NA & -11 &  -11 \\
$A_{7}$ & 27 & \bf{-13} & -13 & -12 &  -12 \\
$A_{8}$ & 27 & \bf{-4} & -4 & -4 &  -4 \\
$A_{9}$ & 27 & \bf{-7} & -7 & -7 &  -7 \\
$A_{10}$ & 27 & \bf{-11} & NA & -11 &  -11\\
\bottomrule
\end{tabular}
\end{sc}
\end{small}
\end{center}
\end{table}

\begin{figure}[ht]
\vskip 0.2in
\begin{center}
\begin{subfigure}{0.48\columnwidth}
\centerline{\includegraphics[width=\columnwidth]{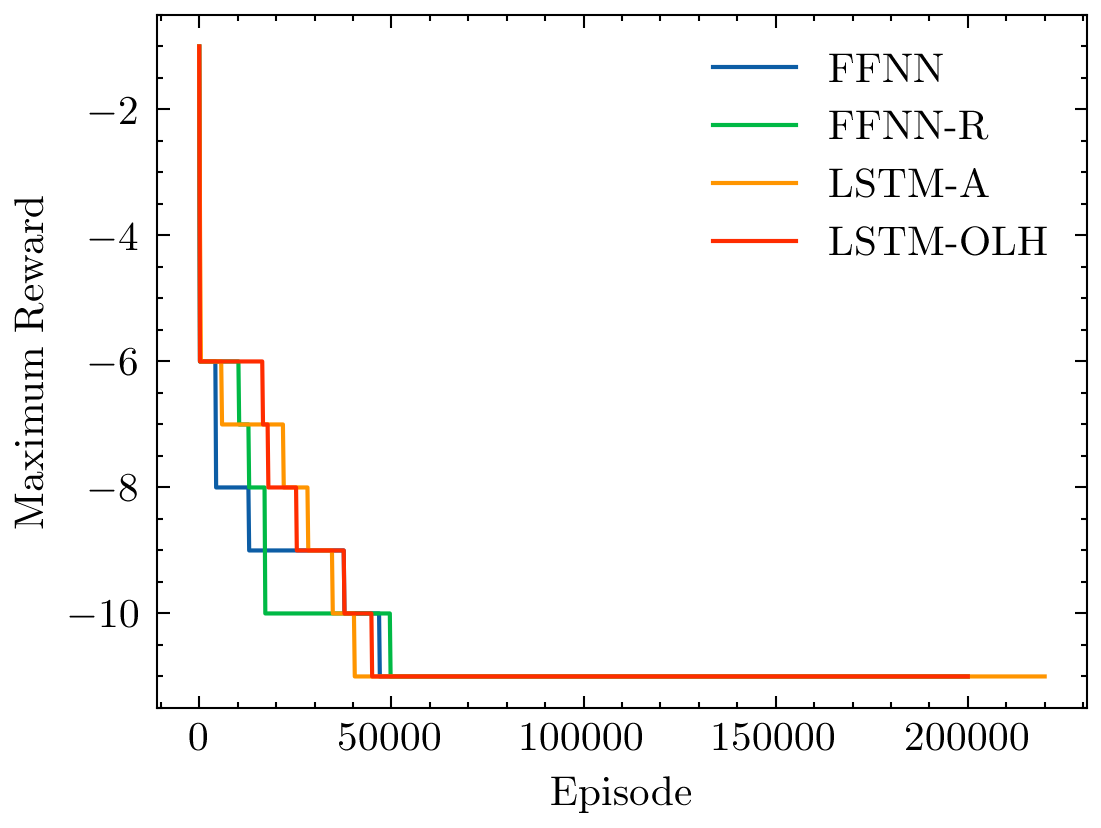}}
\caption{3d1}
\label{fig:sub1}
\end{subfigure}
\begin{subfigure}{0.48\columnwidth}
\centerline{\includegraphics[width=\columnwidth]{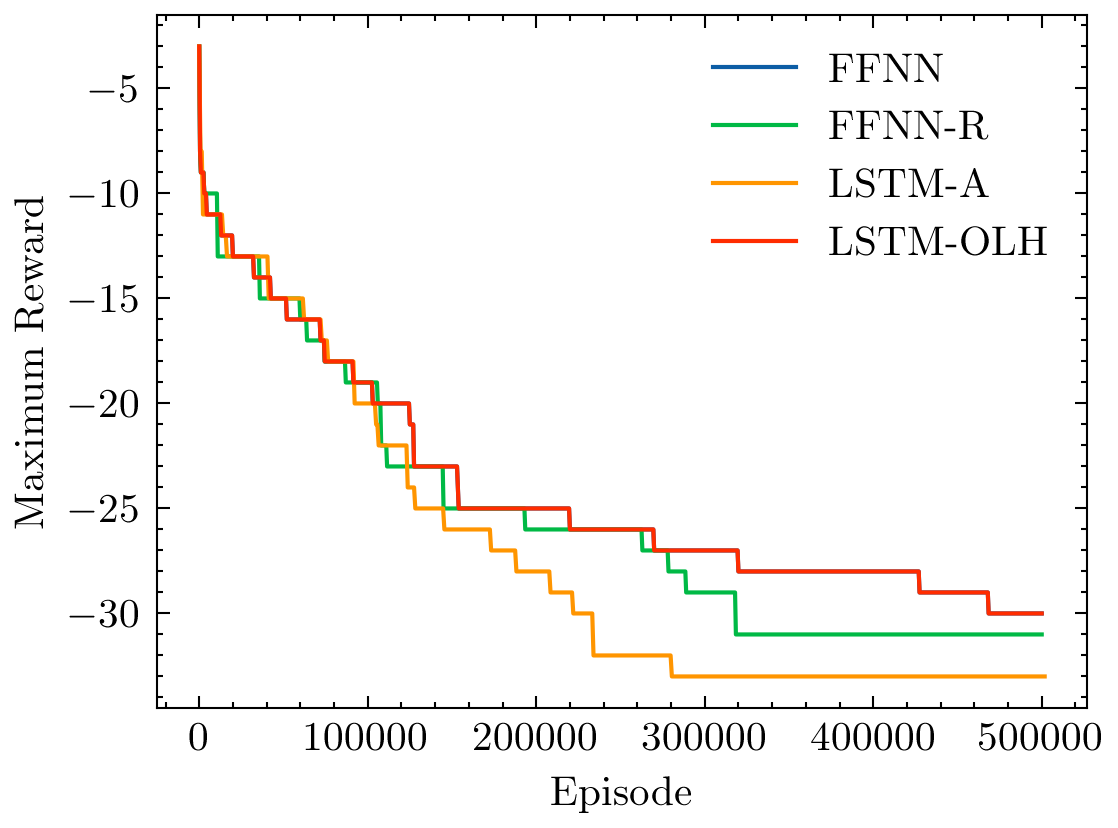}}
\caption{3d5}
\label{fig:sub2}
\end{subfigure}
\caption{Plots a) 3d1 and b) 3d5 show the minimum conformation energy as a function of episode.}
\label{fig:test}
\end{center}
\end{figure}

\begin{figure}[ht]
\vskip 0.2in
\begin{center}
\begin{subfigure}{0.48\columnwidth}
\centerline{\includegraphics[width=\columnwidth]{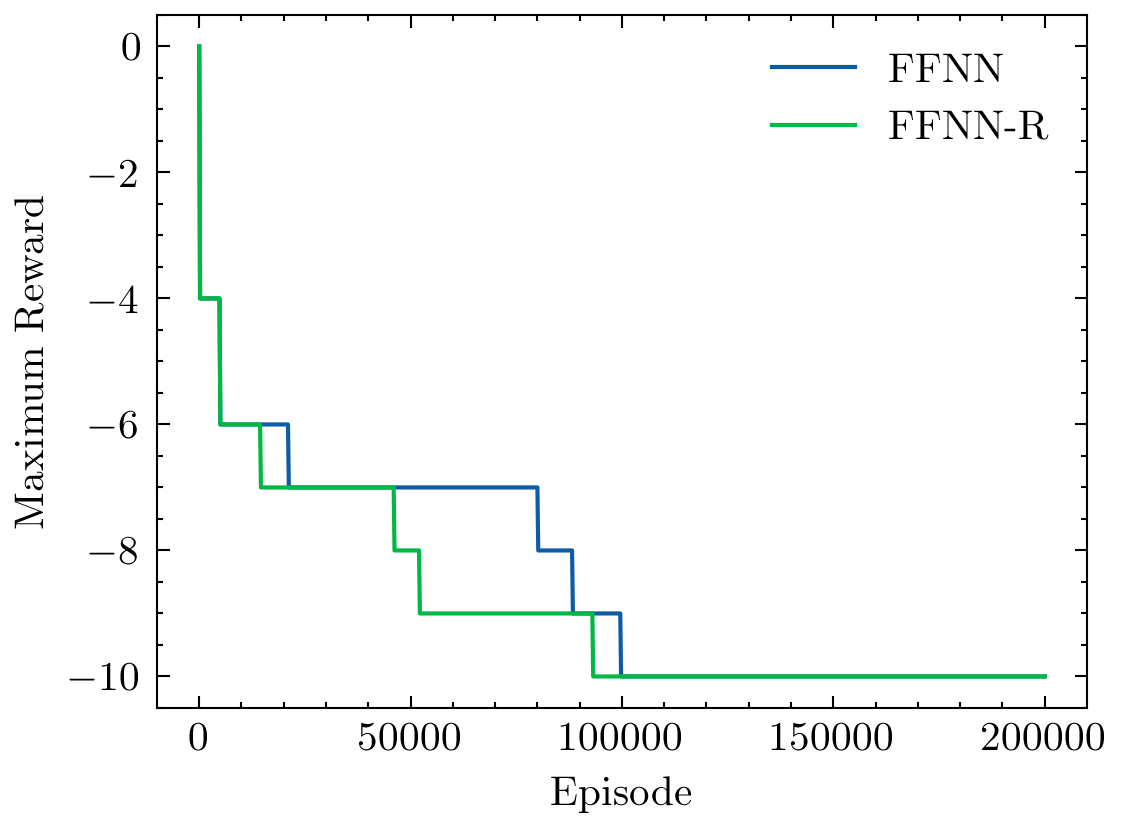}}
\caption{A2}
\label{fig:sub1}
\end{subfigure}
\begin{subfigure}{0.48\columnwidth}
\centerline{\includegraphics[width=\columnwidth]{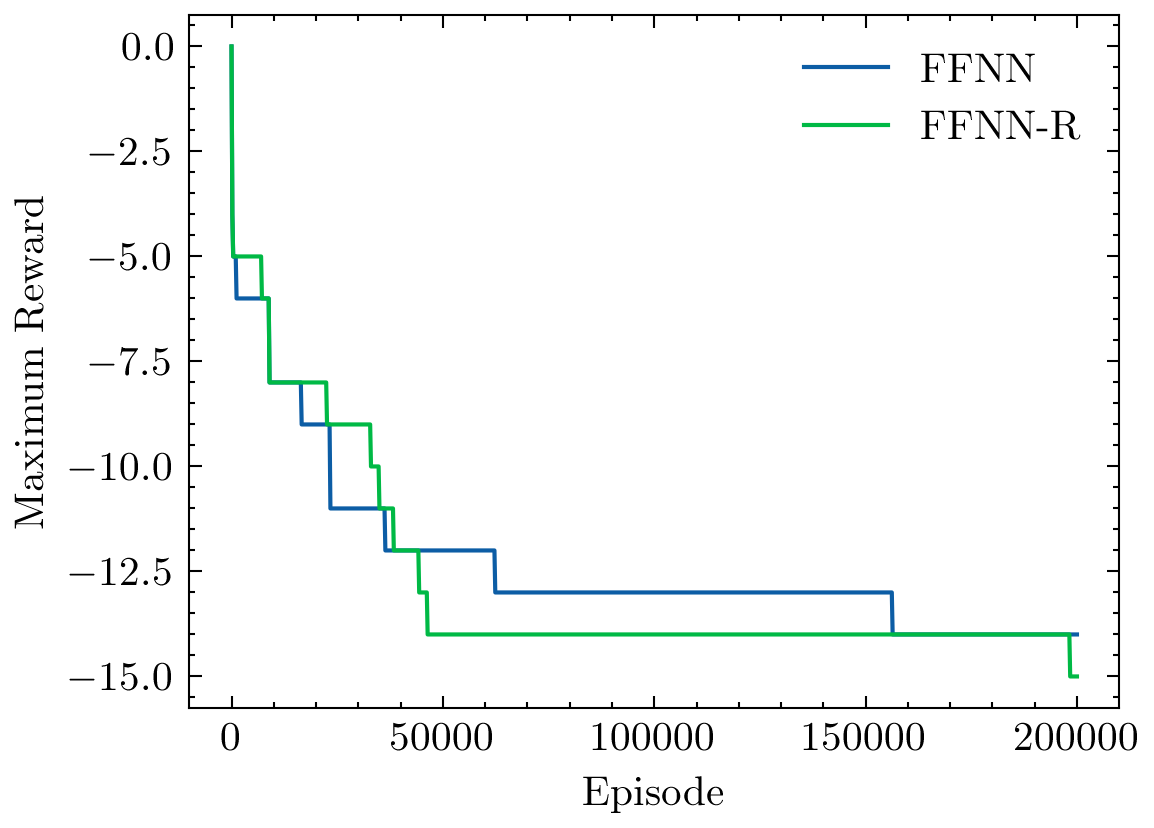}}
\caption{A4}
\label{fig:sub2}
\end{subfigure}
\caption{Plots a) A2 and b) A4 show the minimum conformation energy as a function of episode.}
\label{fig:test}
\end{center}
\end{figure}

\subsection{Efficiency}

The LSTM-A and FFNN-R architectures demonstrate distinct computational profiles in memory usage and training efficiency. LSTM-A requires 2.4GB memory for sequences $\le$ 36 residues (3d1-3d4), with memory distributed across LSTM layers (512 units, 1.8GB), multi-head attention mechanism (0.4GB), and auxiliary layers (0.2GB). FFNN-R maintains a consistent 800MB footprint, comprising a sparse reservoir (1000 nodes, 10\% connectivity, 400KB), input weights (1000 × 8, 32KB), and fully connected layers (256 and 84 nodes, ~1MB), with remaining memory allocated to batch processing and gradients.
Training performance reveals significant differences between architectures. LSTM-A requires 500,000 episodes for optimal convergence on longer sequences (48-50 residues), with training times averaging 8 hours for sequences $\le$ 36 and extending to 48 hours for sequences 3d5-3d9. In contrast, FFNN-R achieves convergence in approximately 200,000 episodes for shorter sequences ($\le$ 36 residues), completing training in 1.5 hours.
FFNN-R's reservoir demonstrates efficient exploration, achieving 80-90\% unique conformational states within the first 50,000 episodes before focusing on promising conformations. LSTM-A compensates for its higher computational demands through parallel processing via its multi-head attention mechanism, enabling efficient handling of longer sequences while reducing computational bottlenecks.

\begin{figure}[ht]
\vskip 0.2in
\begin{center}
\centerline{\includegraphics[scale=0.5]{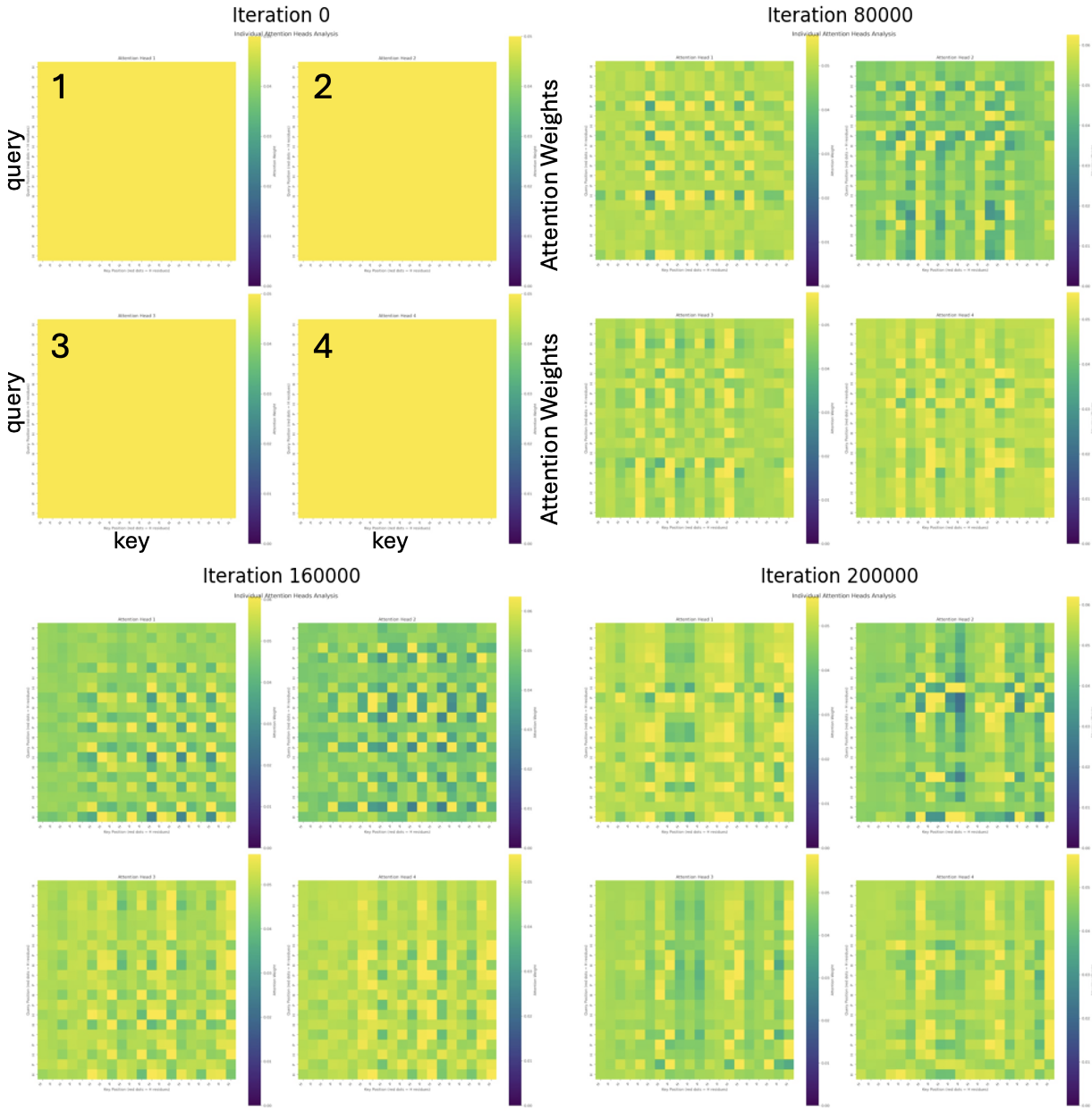}}
\caption{ Evolution of multi-head attention weights during protein structure prediction training. Each row shows the attention patterns of four attention heads at different training iterations (0, 80,000, 160,000, and 200,000). The x and y axes represent query and key positions in the protein sequence, with colors indicating attention strength (purple: 0.0 to yellow: 0.05).}
\label{training_process}
\end{center}
\end{figure}

\section{Discussion}
Our analysis reveals several key insights into both architectures' effectiveness. The LSTM-A's 4-head attention mechanism proves optimal for capturing long-range dependencies in protein folding, as evidenced by the attention weight patterns shown in Figure 7. The visualization demonstrates how different attention heads specialize over training iterations -- initially showing uniform weights (yellow matrices) that evolve into distinct patterns capturing both local and global protein structure interactions. Network depth beyond 5 layers shows a decrease in performance, while batch size of 32 provides optimal balance between memory usage and training stability. \\ Compared to traditional genetic algorithms and Monte Carlo methods, the LSTM-A demonstrates superior performance on longer sequences while requiring significantly less computational time to reach optimal conformations. The evolution of attention weights in Figure 7 reveals how Head 1 develops periodic patterns suggesting secondary structure detection, while Head 2 shows strong position-specific interactions through deeper blue-green regions. Heads 3 and 4 capture broader contextual patterns with more diffuse attention distribution, enabling the model to simultaneously process both local and long-range amino acid interactions. \\ The FFNN-R's success relies on its 1000-neuron reservoir for sequences up to length 36, with linear scaling required for longer sequences. The reservoir's ability to implicitly model temporal dependencies allows the network to capture essential folding patterns without explicit recurrent connections, contributing to its computational efficiency. When compared to traditional FFNN and hybrid approaches like BILS, the FFNN-R shows faster convergence and better energy minimization for shorter sequences. Both FFNN and FFNN-R demonstrate robust performance across multiple independent training runs. While traditional approaches like ant colony optimization and pruned-enriched Rosenbluth method often require multiple restarts to achieve optimal results, our architectures show consistent performance with lower variance in final energy values. \\ However, several important limitations emerge at scale. LSTM-A training becomes computationally demanding beyond 60 amino acids, while FFNN-R performance degrades significantly after length 36. Memory requirements for LSTM-A scale quadratically with sequence length, potentially limiting application to very long sequences. The performance on benchmark sequences suggests both architectures capture fundamental principles of protein folding within the HP model. The LSTM-A's success on longer sequences indicates effective modeling of the hierarchical nature of protein folding, where local structures form first and then assemble into global conformations, as demonstrated by the progressive specialization of attention heads in Figure 7. The FFNN-R's efficiency on shorter sequences suggests rapid learning of local interactions, often sufficient for determining smaller protein structures, providing a more efficient alternative to traditional Monte Carlo sampling approaches. \\ When compared to core-directed chain growth methods and hybrid evolutionary algorithms, both LSTM-A and FFNN-R show comparable or superior performance in terms of final energy values while requiring less parameter tuning and fewer computational resources. The attention mechanism in LSTM-A particularly excels at capturing the kind of long-range interactions that traditional methods often struggle to model effectively, with the attention weights visualization in Figure 7 showing clear evidence of the model learning to focus on both local structural motifs and distant amino acid interactions during training.

\subsection{Limitations and Future Directions}
Our implementation achieves state-of-the-art results for the HP model, though important limitations remain. The reservoir-based approach shows decreased performance for proteins exceeding 36 residues, while the LSTM-A architecture, though more robust for longer sequences, demands substantial computational resources and training time.
Future work should address several key challenges: extending the architectures to handle realistic protein force fields, incorporating additional physical constraints, optimizing the attention mechanism's efficiency for longer sequences, and exploring hybrid approaches that leverage the strengths of both architectures. The linear scaling relationship between sequence length and reservoir size also merits investigation for improved efficiency. The HP model provides an ideal benchmark system, capturing fundamental protein folding aspects while remaining computationally tractable. Its NP-complete nature in finding optimal conformations makes it particularly valuable for testing new algorithms. Notably, our methods successfully achieve optimal conformations that align with known energy minima.

\section{Conclusion}

This study advances protein structure prediction through two novel architectures. The LSTM-A achieves good performance on longer sequences (N $>$ 36), while the FFNN-R provides efficient solutions for shorter sequences. Both consistently match best-known energy values across benchmark sequences. The success of these architectures demonstrates the viability of deep learning approaches in protein structure prediction, while establishing new benchmarks for computational efficiency. Their complementary strengths suggest promising directions for handling proteins of varying lengths and complexities. Future work should explore hybrid architectures combining LSTM-A and FFNN-R strengths, more efficient attention mechanisms for longer sequences, and parallel training strategies. 

\section*{Software and Data}
The software and data used in this study are available in a public \href{https://github.com/AnonymousAuthor117/Protein-Structure-Prediction-in-the-3D-HP-Model-Using-Deep-Reinforcement-Learning}{GitHub} repository. This repository contains the implementation of our protein structure prediction model using deep reinforcement learning in the 3D HP model.

\section*{Acknowledgements}
This work was supported by the National Institutes of Health (R01-GM148586). Simulations were run on the Hive cluster, which is supported by the National Science Foundation under grant number 1828187, and the Phoenix cluster, both of which are managed by the Partnership for an Advanced Computing Environment
(PACE) at the Georgia Institute of Technology.

\bibliography{references}

\appendix
\renewcommand{\thetable}{A.\arabic{table}} % Change table numbering to A.1, A.2, etc.
\setcounter{table}{0} % Reset table counter

\section{Experiments}
All of the conducted experiments are tabulated in Table A.1.

\begin{landscape}
\begin{table}[htbp]
\centering
\caption{3D Trial Runs, Dictionary: LSTM-OLH → LSTM - OnlyLastHidden, LSTM-A (Num of Heads) → LSTM - with Attention, FNN-VANILLA → Without Reservoir, FNN-Reservoir → With Reservoir}
\adjustbox{max width=1.45\textwidth}{  % Adjust this value as needed
\begin{tabular}{|l|l|c|c|c|c|c|c|c|c|c|}
\hline
\textbf{Sequence} & \textbf{Architecture} & \textbf{Num Layers} & \textbf{Size of Layer} & \textbf{Batch Size} & \textbf{Num Training Episodes} & \textbf{Time to Train} & \textbf{Date} & \textbf{E\_Min Best} & \textbf{E\_min Experimental} & \textbf{Num Parameters} \\
\hline
3d1 & LSTM-A (4) & 3 & 512 & 16 & 200K & 6:54:38 & 10/12/24 & -11 & -11 & 6324741 \\
3d2 & LSTM-A (4) & 3 & 512 & 16 & 220K & 7:33:33 & 10/12/24 & -13 & -13 & 6324741 \\
3d3 & LSTM-A (4) & 3 & 512 & 16 & 220K & 7:54:00 & 12/12/24 & -9 & -9 & 6324741 \\
3d4 & LSTM-A (4) & 3 & 512 & 16 & 250K & 11:14:39 & 10/12/24 & -18 & -18 & 6324741 \\
3d5 & LSTM-A (4) & 5 & 512 & 16 & 500K & 41:27:15 & 10/13/24 & -35 & -33 & 10527237 \\
3d6 & LSTM-A (4) & 5 & 512 & 16 & 500K & 41:23:16 & 10/13/24 & -31 & -30 & 10527237 \\
3d5 & LSTM-A (4) & 5 & 512 & 32 & 750K & 45:15:55 & 10/20/24 & -35 & -33 & 10527237 \\
3d7 & LSTM-A (4) & 5 & 512 & 32 & 750K & 49:49:49 & 10/20/24 & -34 & -32 & 10527237 \\
3d6 & LSTM-A (4) & 5 & 512 & 32 & 750K & 46:15:23 & 10/22/24 & -31 & -30 & 10527237 \\
3d8 & LSTM-A (4) & 5 & 512 & 32 & 750K & 57:00:00 & 10/22/24 & -44 & -40 & 10527237 \\
3d8 & LSTM-A (4) & 6 & 512 & 32 & 750K & - & 10/22/24 & -44 & - & - \\
3d9 & LSTM-A (4) & 5 & 512 & 32 & 750K & 65:00:00 & 10/26/24 & -55 & -51 & -\\
3d1 & LSTM-OLH & 3 & 512 & 16 & 200K & 3:18:16 & 10/26/24 & -11 & -11 & 5274117 \\
3d2 & LSTM-OLH & 3 & 512 & 16 & 220K & 4:06:36 & 10/26/24 & -13 & -13 & 5274117 \\
3d3 & LSTM-OLH & 3 & 512 & 16 & 220K & 1:06:10 & 10/26/24 & -9 & -9 & 5274117 \\
3d4 & LSTM-OLH & 3 & 512 & 16 & 250K & 1:24:51 & 10/26/24 & -18 & -18 & -\\
3d5 & LSTM-OLH & 5 & 512 & 16 & 500K & - & 10/26/24 & -35 & -35 & -\\
3d6 & LSTM-OLH & 5 & 512 & 16 & 500K & - & 10/26/24 & -  & - & -\\
3d7 & LSTM-OLH & 5 & 512 & 32 & 750K & - & 10/26/24 & - & - & -\\
3d8 & LSTM-OLH & 5 & 512 & 32 & 750K & - & 10/26/24 & - & - & -\\
3d9 & LSTM-OLH & 5 & 512 & 32 & 750K & - & 10/26/24 & - & - & -\\
3d1 & FNN-VANILLA & 4 & 512, 256, 84 & 16 & 200K & 0:58:15 & 10/26/24 & -11 & -11 & -\\
3d2 & FNN-VANILLA & 4 & 512, 256, 84 & 16 & 220K & 1:06:16 & 10/26/24 & -13 & -13 & -\\
3d3 & FNN-VANILLA & 4 & 512, 256, 84 & 16 & 220K & 1:06:29 & 10/26/24 & -9 & -9 & -\\
3d4 & FNN-VANILLA & 4 & 512, 256, 84 & 16 & 250K & 1:25:10 & 10/26/24 & -18 & -18 & -\\
3d5 & FNN-VANILLA & 4 & 512, 256, 84 & 16 & 500K & 3:13:46 & 10/26/24 & -35 & -30 & -\\
A1 & FNN-VANILLA & 4 & 512, 256, 84 & 16 & 200K & 1:08:39 & 10/30/24 & -9 & -9 & -\\
A2 & FNN-VANILLA & 4 & 512, 256, 84 & 16 & 200K & 1:08:18 & 10/30/24 & -10 & -10 & -\\
A3 & FNN-VANILLA & 4 & 512, 256, 84 & 16 & 200K & 1:26:09 & 10/30/24 & -8 & -8 & -\\
A4 & FNN-VANILLA & 4 & 512, 256, 84 & 16 & 200K & 1:27:57 & 10/30/24 & -15 & -14 & -\\
A5 & FNN-VANILLA & 4 & 512, 256, 84 & 16 & 200K & - & 10/30/24 & -8 & -8 & -\\
A6 & FNN-VANILLA & 4 & 512, 256, 84 & 16 & 200K & 1:09:17 & 10/30/24 & -12 & -11 & -\\
A7 & FNN-VANILLA & 4 & 512, 256, 84 & 16 & 200K & 1:27:20 & 10/30/24 & -13 & -12 & -\\
A8 & FNN-VANILLA & 4 & 512, 256, 84 & 16 & 200K & 1:27:26 & 10/30/24 & -4 & -4 & -\\
A9 & FNN-VANILLA & 4 & 512, 256, 84 & 16 & 200K & 1:27:50 & 10/30/24 & -7 & -6 & -\\
A10 & FNN-VANILLA & 4 & 512, 256, 84 & 16 & 200K & 1:27:57 & 10/30/24 & -11 & -11 & -\\
A1 & FNN-RESERVOIR & 4 & 512, 256, 84 & 16 & 200K & 1:12:23 & 11/2/2024 & -9 & -9 & 264445 \\
A2 & FNN-RESERVOIR & 4 & 512, 256, 84 & 16 & 200K & 1:27:59 & 11/2/2024 & -10 & -10 & -\\
A3 & FNN-RESERVOIR & 4 & 512, 256, 84 & 16 & 200K & 1:13:31 & 11/2/2024 & -8 & -8 & -\\
A4 & FNN-RESERVOIR & 4 & 512, 256, 84 & 16 & 200K & 1:14:01 & 11/2/2024 & -15 & -14 & -\\
A5 & FNN-RESERVOIR & 4 & 512, 256, 84 & 16 & 200K & - & 11/2/2024 & -8 & - & -\\
A6 & FNN-RESERVOIR & 4 & 512, 256, 84 & 16 & 200K & 1:27:23 & 11/2/2024 & -12 & -11 & -\\
A7 & FNN-RESERVOIR & 4 & 512, 256, 84 & 16 & 200K & 1:28:23 & 11/2/2024 & -13 & -12 & -\\
A8 & FNN-RESERVOIR & 4 & 512, 256, 84 & 16 & 200K & 1:27:01 & 11/2/2024 & -4 & -4 & -\\
A9 & FNN-RESERVOIR & 4 & 512, 256, 84 & 16 & 200K & 1:27:33 & 11/2/2024 & -7 & -6 & -\\
A10 & FNN-RESERVOIR & 4 & 512, 256, 84 & 16 & 200K & 1:09:38 & 11/2/2024 & -11 & -11 & -\\
\hline
\end{tabular}
}
\end{table}
\end{landscape}

\newpage

\section{Additional Figures}

\renewcommand{\thefigure}{B.\arabic{figure}} % Change figure numbering to A.1, A.2, etc.
\setcounter{figure}{0} % Reset figure counter

\begin{figure*}[h]
    \centering
    \includegraphics[scale=0.4]{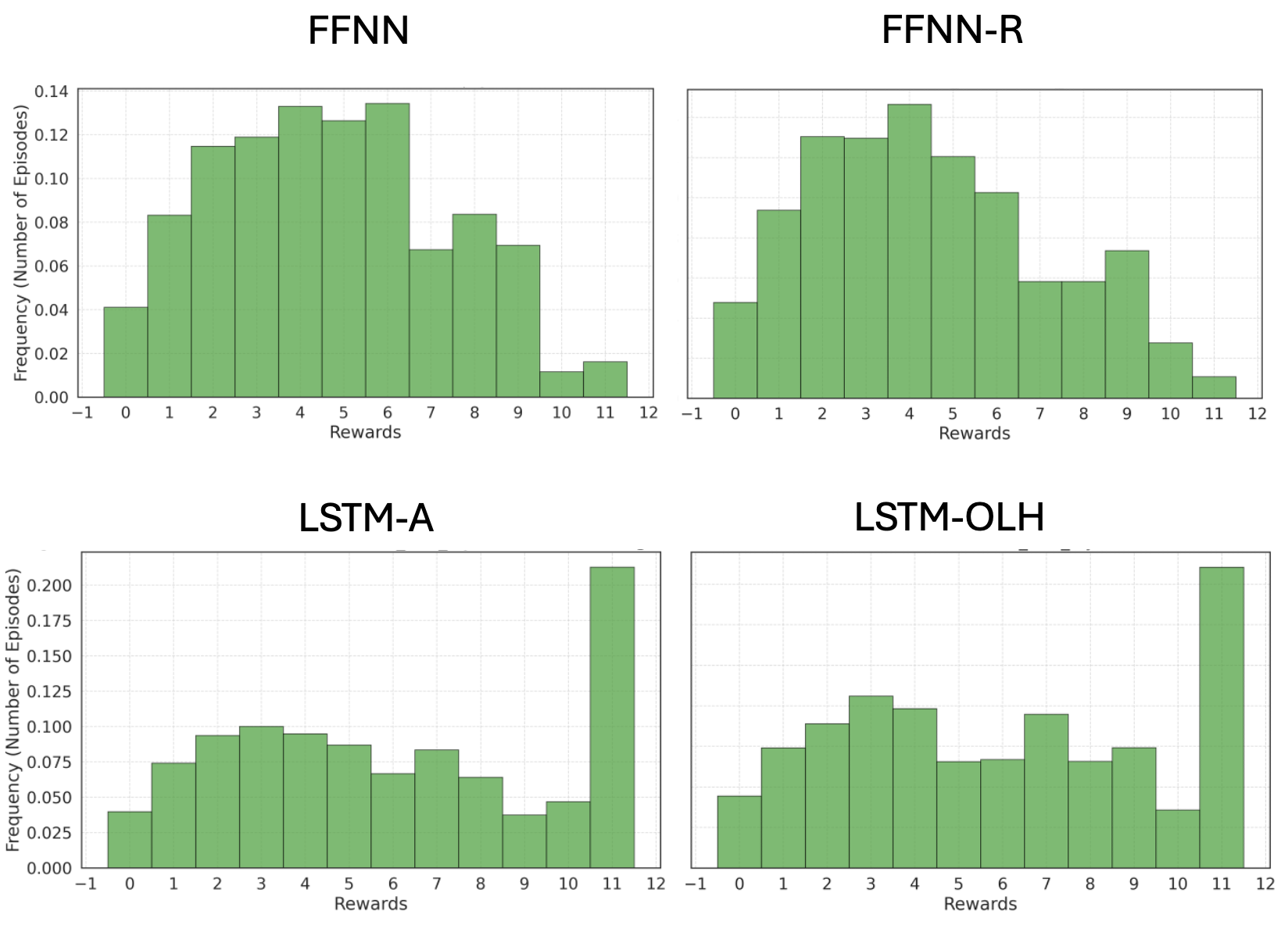}
    \caption{Reward histograms across the four studied architectures when training for the 3d1 sequence.}
    \label{fig:enter-label}
\end{figure*}

\begin{figure}[h]
    \centering
    \includegraphics[scale=0.4]{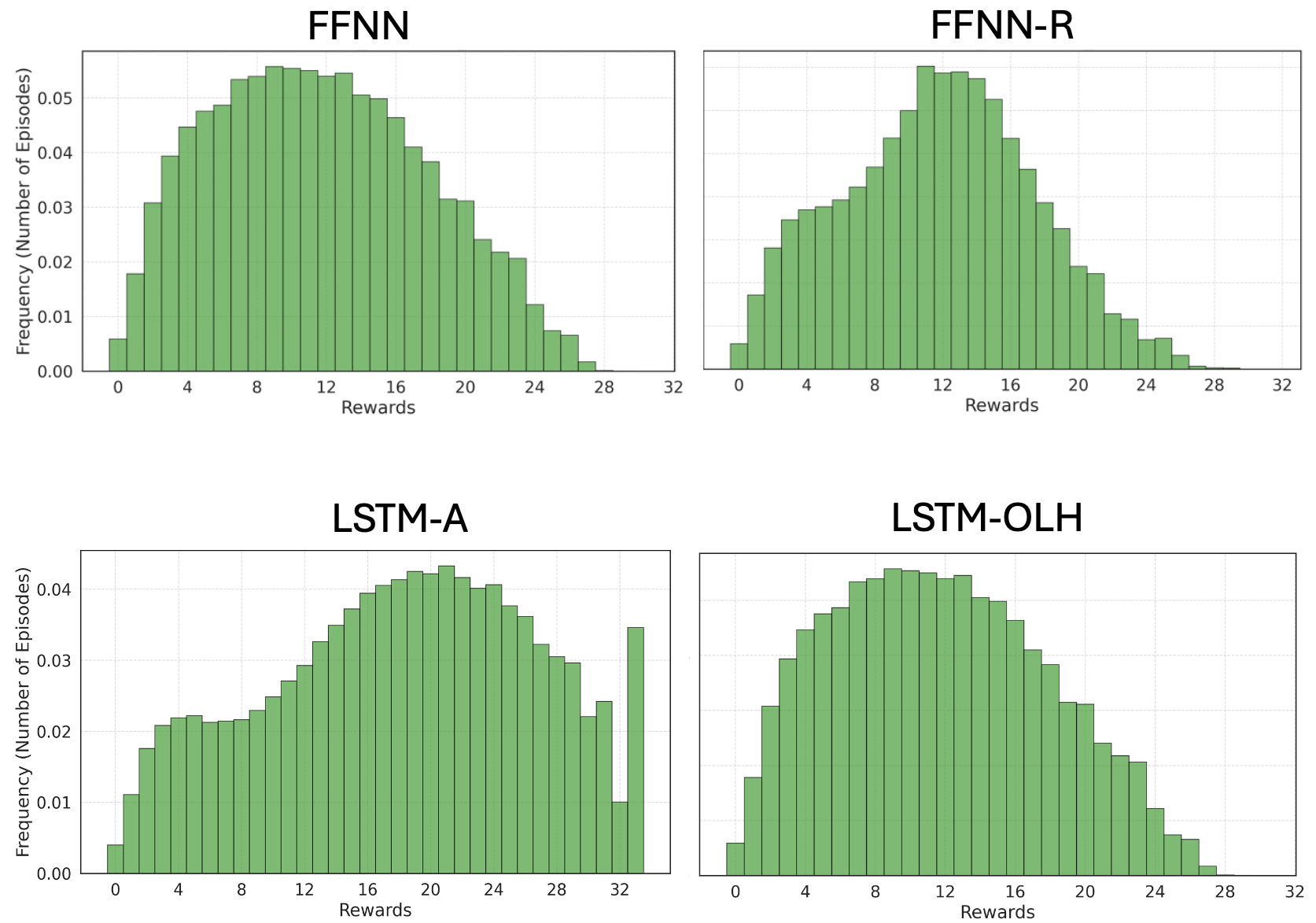}
    \caption{Reward histograms across the four studied architectures when training for the 3d5 sequence.}
    \label{fig:enter-label}
\end{figure}

\begin{figure*}[h]
    \centering
    \includegraphics[scale=0.4]{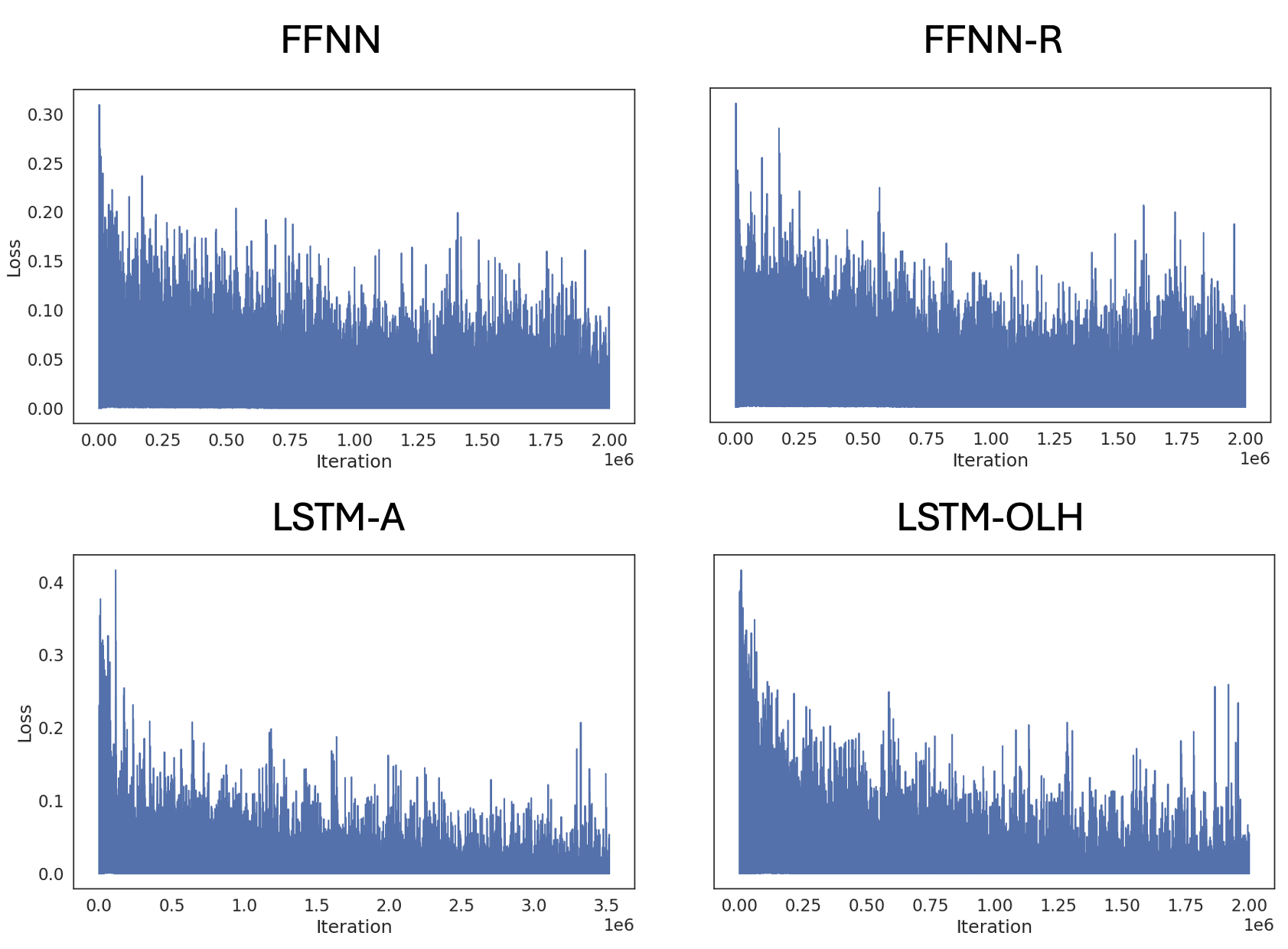}
    \caption{Loss function across the four studied architectures when training for the 3d1 sequence.}
    \label{fig:enter-label}
\end{figure*}

\begin{figure}[h]
    \centering
    \includegraphics[scale=0.4]{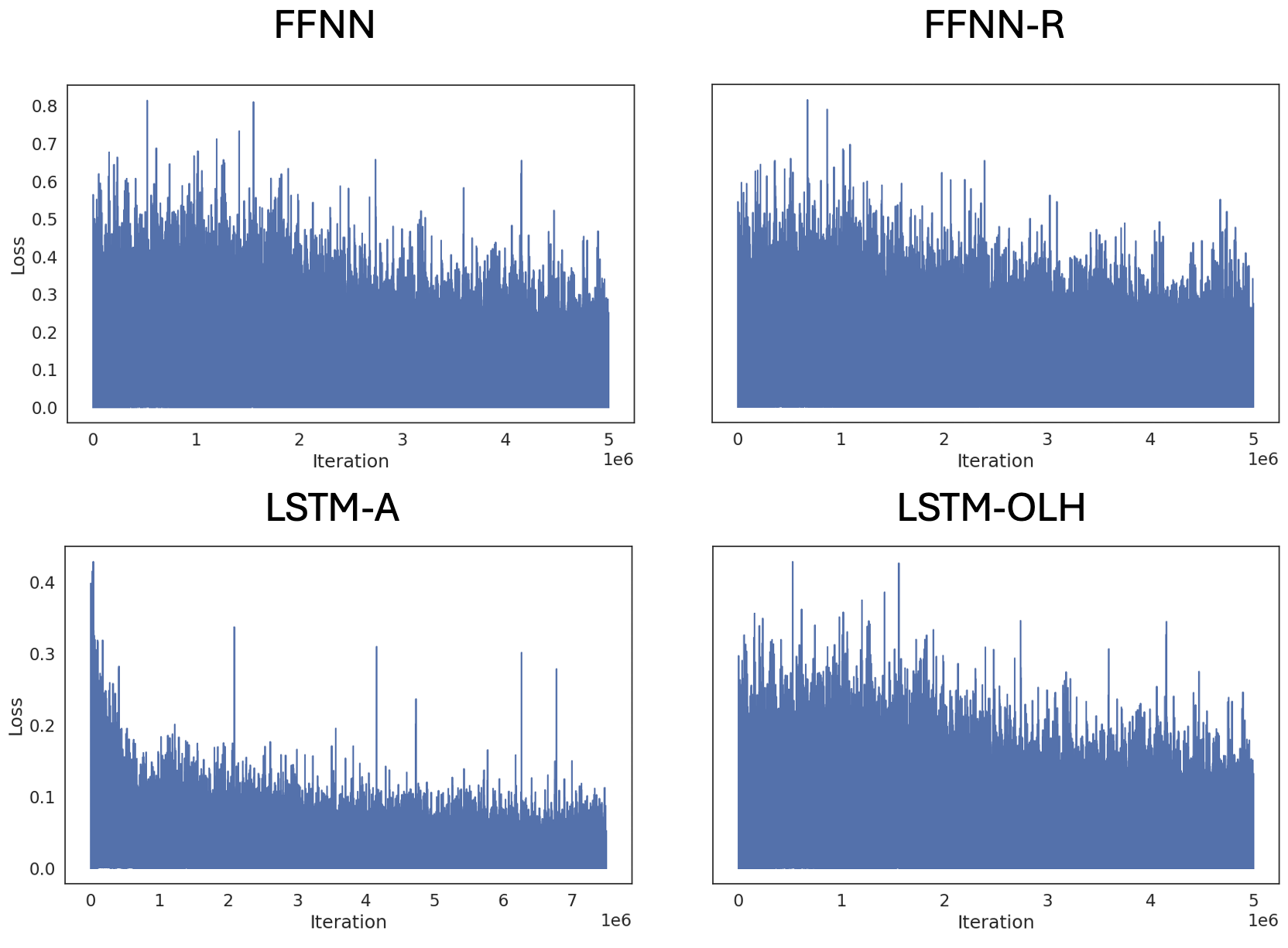}
    \caption{Loss function across the four studied architectures when training for the 3d5 sequence.}
    \label{fig:enter-label}
\end{figure}

\end{document}